%% file: main.tex
\documentclass[sigconf]{acmart}
\usepackage{bm}
\usepackage{algorithm}
\usepackage{multirow}
\usepackage[noend]{algorithmic}
\usepackage{subfigure}
\usepackage{tikz}
\usepackage{url}
\usepackage{xspace}
\newcommand*{\circled}[1]{\lower.7ex\hbox{\tikz\draw (0pt, 0pt)circle (.5em) node {\makebox[1em][c]{\small #1}};}}

\usepackage{wrapfig}
\usepackage{subfigure}
\usepackage{multirow}
\usepackage{bm}
\usepackage{bbm}
\usepackage{color, soul}
\usepackage{subfigure}
\usepackage{graphicx}
\usepackage{listings}
\usepackage{xcolor}
\usepackage{makecell}

\usepackage{color, soul}
\usepackage{bm}
\usepackage{CJKutf8}
\usepackage{booktabs}
\usepackage{tabularx}

\newcommand{\red}[1]{\textcolor{red}{#1}}

\usepackage{xcolor}

\newcommand{\para}[1]{{\vspace{5pt} \bf \noindent #1 \hspace{5pt}}}

\newcommand{\sys}{\textsc{IPT}\xspace}

\AtBeginDocument{%
  \providecommand\BibTeX{{%
    \normalfont B\kern-0.5em{\scshape i\kern-0.25em b}\kern-0.8em\TeX}}}

\begin{document}
\fancyhead{}

\title{Instance-wise Prompt Tuning for Pretrained Language Models}

\author{Yuezihan Jiang$^\dagger$, Hao Yang$^{\ddagger}$, Junyang Lin$^{\ddagger}$, Hanyu Zhao$^{\ddagger}$, An Yang$^{\ddagger}$, Chang Zhou$^{\ddagger}$, Hongxia Yang$^{\ddagger}$, Zhi Yang$^{\dagger}$, Bin Cui$^{\dagger \S}$}
\affiliation{
 {{$^\dagger$}{School of EECS \& Key Laboratory of High Confidence Software Technologies, Peking University}}~~~~~
$^\ddagger$Alibaba Group\country{}}
\affiliation{
$^\dagger$\{jiangyuezihan, yangzhi, bin.cui\}@pku.edu.cn\\ $^\ddagger$\{yh351016, junyang.ljy, zhaohanyu, ya235025, ericzhou.zc, yang.yhx\}@alibaba-inc.com\country{}
}

\begin{abstract}
Prompt Learning has recently gained great popularity in bridging the gap between pretraining tasks and various downstream tasks. It freezes Pretrained Language Models (PLMs) and only tunes a few task-related parameters (prompts) for downstream tasks, greatly reducing the cost of tuning giant models. The key enabler of this is the idea of querying PLMs with task-specific knowledge implicated in prompts. This paper reveals a major limitation of existing methods that the indiscriminate prompts for all input data in a task ignore the intrinsic knowledge from input data, resulting in  sub-optimal performance. 

We introduce \emph{Instance-wise Prompt Tuning (\sys)}, the first prompt learning paradigm that injects knowledge from the input data instances to the prompts, thereby providing PLMs with richer and more concrete context information. We devise a series of strategies to produce instance-wise prompts, addressing various concerns like model quality and cost-efficiency. Across multiple tasks and resource settings, IPT significantly outperforms task-based prompt learning methods, and achieves comparable performance to conventional finetuning with only 0.5\% - 1.5\% of tuned parameters.
\end{abstract}



\maketitle

\input{1.intro}
\input{2.pre}

\input{3-4}

\input{4.method}
\input{5.experiment}

\section{Conclusion}
\label{conclusion}
In this paper, we propose a new prompt tuning paradigm called Instance-wise Prompt Tuning (\sys). Different from previous works which utilize task-specific knowledge to query PLMs, \sys provide instance-wise prompt strategies to boost downstream tasks by querying PLMs with data intrinsic knowledge. Furthermore, the results on a wide range of NLU tasks under different resource limitations suggest that, instance-wise prompts can provide efficient and effective solutions to leverage PLMs in a lightweight manner. Besides, it is worthy of investigating how to combine previous task-based methods and the instance-wise prompt method together, and we leave this in future works. 





\bibliographystyle{ACM-Reference-Format}
\bibliography{my-reference}

\clearpage
\input{6.appendix}

\end{document}

%% file: 1.intro.tex
\section{Introduction}
In these years, we have witnessed the tremendous success of large-scale pretraining on natural language~\cite{bert, gpt, gpt3}. Pretrained language models (PLMs) on large-scale data have promoted  achievements in a series of downstream tasks, including natural language understanding (NLU)~\cite{glue, wang2019superglue}, natural language generation (NLG)~\cite{unilm, bart}, etc. This attributes to the pretraining-finetuning paradigm, where the model is pretrained on large-scale unsupervised data and finetuned on task-specific data. However, finetuning on all model parameters is essentially memory-inefficient and requires copies of models for different downstream tasks. 





Prompt Learning \cite{gpt3,li2021prefix, lester2021power, liu2021p} has been an appealing approach to making large PLMs more cost-efficient. The intuition is that a PLM has accommodated a massive amount of knowledge; the key of a downstream task is to retrieve the needed part. To this end, prompt learning equips PLMs with certain ``prompts'' that suggest a specific context when tuning the model, e.g., domain of the downstream task (\cite{li2021prefix, lester2021power, liu2021p}). Such information helps stimulate  more context-relevant part of the knowledge in a PLM. In this way, people can freeze the PLM parameters and only tune the prompts, thus reducing the cost of tuning a giant model. 
These methods have been shown effective in improving various downstream tasks by combining task-specific knowledge during tuning.

Unfortunately, we find a major limitation of task-based prompts that the prompts are too coarse-grained and \emph{ignore fine-grained information in input data}. In state-of-the-art methods, all input data within a tuning task shares an identical prompt implying the task domain. However, besides the task, the specific content in the input data also contains context that can help the PLM retrieve more relevant knowledge, for example, the specific object being talked about. Such knowledge behinds the input data should be fully exploited to unleash the potential of prompts.

To capture such fine-grained knowledge behind the input data, this paper introduces \emph{Instance-wise Prompt Tuning (IPT)}, a new Prompt Learning paradigm that incorporates the inner knowledge of input data when querying PLMs for downstream tasks, instead of using input-agnostic prompts. The core of IPT is a module called the \emph{IPT strategy}, which is responsible for producing a prompt head for \emph{each instance} of the input data, thereby injecting data-specific knowledge to the PLM. 

We present three different IPT strategies designed for various goals and from various perspectives, including a basic one and two advanced versions. The basic IPT strategy simply uses embeddings of the input words to produce the instance-wise prompts. It involves a prompt table for each word in the vocabulary, which is initialized randomly and trained on the input dataset during tuning. Despite being very straightforward, this strategy can already outperform task-based prompts, showing the effectiveness of IPT.

Our second strategy further enhances the embebdding  of the prompt table by \emph{pretraining with data knowledge}. Training the embeddings only on the input dataset of the downstream task might be insufficient to extract enough background knowledge of the words for the prompts. We hence propose to pretrain the embeddings with extra corpus to acquire richer knowledge. Recall that existing methods use prompts to imply the domains of the downstream tasks and to retrieve the knowledge of relevant categories. Following this intuition, we define a simple yet effective pretraining task that classifies the instance into related categories, so that the input data can query the PLM with the categories to extract its corresponding domain knowledge.
We show that this method requires very little manual labeling efforts but shows great performance improvement over randomly initialized prompts.

Our third strategy shows how to incorporate the \emph{encoder} mechanism based on the pretrained prompts to reduce training costs. Instead of using the embeddings in a prompt table as the prompts, we introduce an encoder neural network to transform the embeddings to latent representations. This enables us to freeze the pretrained prompt table and only optimize the encoder, whose parameters can be smaller than the prompt table, thus requiring a smaller training cost without degrading accuracy.

All of the above combined, IPT achieves three important advantages over existing methods.
Firstly, it is \emph{efficient, in terms of both performance and cost}. By injecting input-specific knowledge into the PLM with carefully designed strategies, IPT outperforms finetuning with only 0.5\% tuned parameters of finetuning, and one-third of prefix tuning. Second, it is \emph{universally applicable}. The benefits of prior methods are mostly restricted in certain scenarios, but IPT relaxes some constraints (e.g., large-scale PLMs, few/full-shot settings) and shows great performance consistently. Finally, it is \emph{general and flexible}. IPT 
is a general framework compatible with existing Prompt Learning pipelines and allows various IPT strategy designs. It also allows customization within specific strategies, e.g., pretraining methods or encoder types. 
IPT opens up  new design space for future research to explore more design choices.

To summarize, the contributions of the paper are listed below:
\begin{itemize}
    \item \emph{The IPT concept.} To the best of our knowledge, we are the first to propose the concept and workflow of instance-wise prompt tuning, which injects input-specific knowledge for prompt tuning to the pretrained model. 
    \item \emph{Knowledge-enhanced pretraining.} We develop a simple and effective data-based pretraining strategy that enhances the prompts with more domain knowledge of the data.
    \item \emph{Encoders for reducing training costs.} We incorporate encoders to transform pretrained embeddings, facilitating freezing the embeddings and only tuning the encoders.
    \item \emph{Experimental evaluation.} Comprehensive experiments confirm the efficiency and universality of IPT.
    
\end{itemize}


%% file: 2.pre.tex
\section{Related Work}
\label{related}
\para{Prompt Learning}
Since the first emergence in GPT-3~\cite{vaswani2017attention} in a literal way, prompts show their effectiveness in achieving superior performance with in-context learning in low-resource settings. 
The most natural way to create prompts is to manually create intuitive templates based on human introspection. For example,  ~\cite{knowledgebases} provides manually created cloze templates to probe knowledge in LMs.
PET and its variants~\cite{schick2020exploiting,schick2020s} follow this insight and establish a pattern-verbalizer mode for manual prompt design in a few-shot learning setting.
While the strategy of manually designing templates are intuitive and request  expertise knowledge in prompt design.
To avoid dependence on domain knowledge of manual prompt design,~\cite{shin2020autoprompt,schick2020automatically} create prompts  auto-generated by models, in the format of a sequence of discrete trigger words.
~\cite{attackingnlp} applied a gradient-based search over actual tokens to find a sequence of  tokens that can trigger the underlying pre-trained LM to generate  desired target prediction.

Specifically,\cite{qin2021learning,li2021prefix} develop soft prompts with no literal implication.~\cite{houlsby2019parameter} adds trainable and lightweight modules to frozen PLMs for downstream applications and also achieves decent performance. Continuous prompts  relax the constraint that the embeddings of template words are the embeddings of natural language words and the template is parameterized by the pre-trained LM’s parameters. Instead, template words have their own parameters which can be tuned based on training data from the downstream task. 

Some works extend the effects of prompt strategy to different kinds of tasks, e.g. relation extraction~\cite{han2021ptr,chen2021adaprompt}.~\cite{tsimpoukelli2021multimodal} trains a vision encoder which encodes an image into a sequence of embeddings to generate a prompt, which can be used to  a frozen auto-regressive LM to generate the appropriate caption in few-shot learning for vision-language tasks such as visual question answering etc. 

\para{Prefix tuning methods}
\label{prefix_work}
Prefix tuning refers to a prompt tuning method which adopts prompts as tunable prefix  tokens to the input data for the downstream tuning process. Prefix Tuning~\cite{li2021prefix,liu2021p} and its variants directly prepend a sequence of continuous task-specific vectors to the input data at each layer, while keeping the PLM's parameters frozen. 
Their experimental results demonstrate that such continuous prefix-based learning can behave well in natural language generation (NLG) tasks. Similarly, Prompt Tuning~\cite{lester2021power} only prepends a series of continuous task-specific vectors tokens to the input sentences at the input layer, and tunes these tokens while keeping the PLM fixed. An important difference between the two methods is that Prompt Tuning introduces fewer parameters because it only introduces parameters in the input layers and does not add additional tunable parameters within each layer of the based PLMs. Following these mainstream methods, many works extend the task-based prompts to other formats (e.g., Frozen~\cite{tsimpoukelli2021multimodal} uses representations of images as prefix embeddings). 

\para{Knowledge Enhanced LMs}
Knowledge enhancing becomes a fashion to boost language models in recent years. For example, some works introduce extra side information as a supplementary in the pretraining stage~\cite{zhang2019ernie} and the fine-tuning stage~\cite{guan2020knowledge}. With the advance of prompt-tuning, some works pilot knowledge enhance strategy in prompt-tuning.~\cite{hu2021knowledgeable} propose to incorporate external knowledge into the verbalizer with calibration.~\cite{chen2021knowprompt} introduces KnowPrompt to inject entity and relation knowledge into prompt construction.~\cite{gu2021ppt} first explores to pretrain the learnable continuous prompts with a next-sentence-prediction task to endow prompts with syntactic knowledge. ~\cite{qin2021exploring} pretrains a sub-place of original prompts on hundreds of tasks and achieve considerable performance on both few-shot tasks and zero-shot tasks. Different from previous works that directly introduce knowledge to task-specific training process (e.g., applying human-annotated knowledge to the prompting process or pretraining on a fixed embedding space of prompts), we directly perform knowledge enhancement from the data side by introducing a novel knowledge classification task. The pretraining process is more economic because the pretraining process is simple and it hardly needs human anotations.




%% file: 3-4.tex
\section{Motivation}

\label{motivation}
\begin{figure}[tpb]
    \centering
    \includegraphics[width=0.9\linewidth]{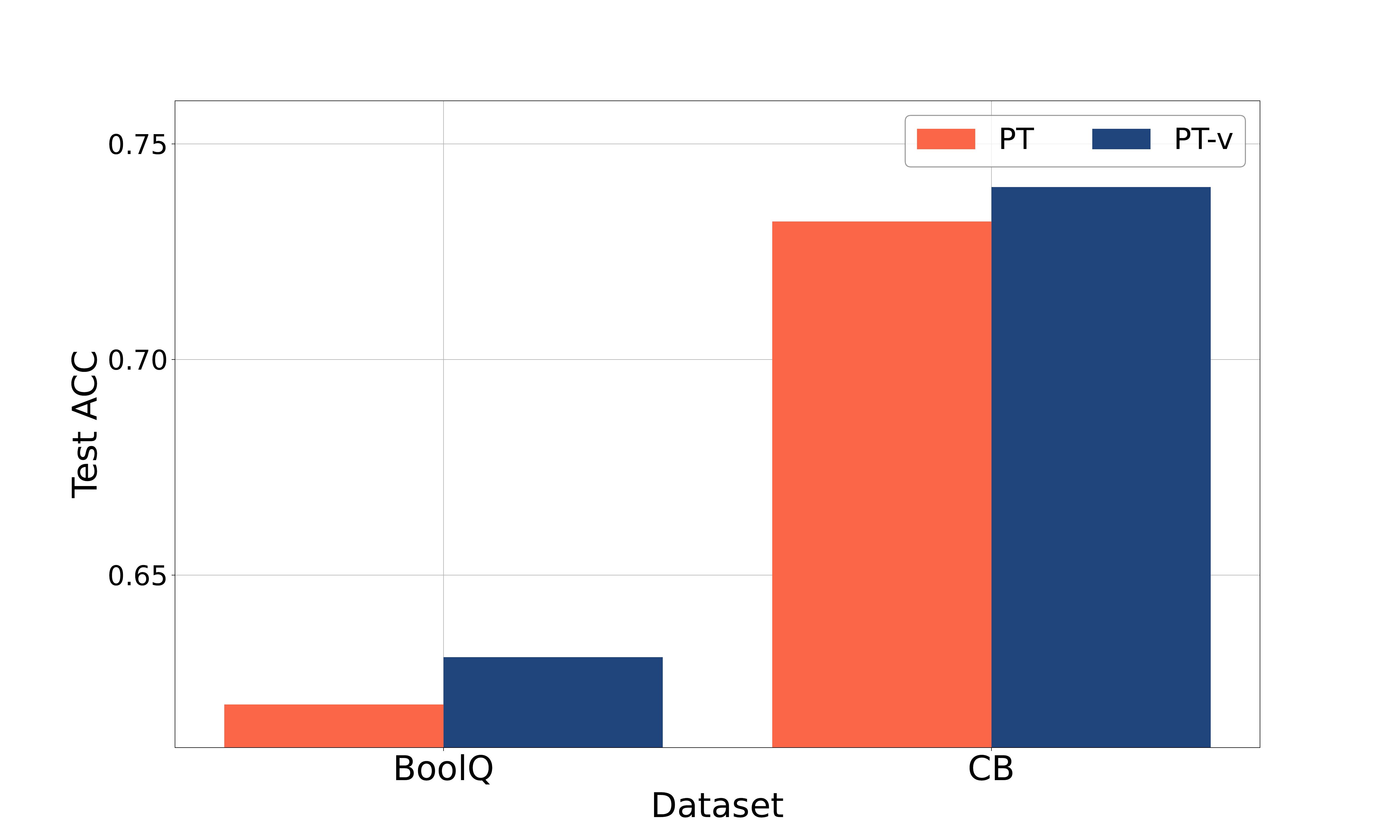}
    \vspace{-4mm}
    \caption{Instance-wise prompts on Prompt Tuning.}
    \label{observation}
    \vspace{-5mm}
\end{figure}

\para{Formulation in Prompt Learning methods.}
\label{formulation_general}
Prompt Learning methods define a sequence of prompt tokens $\mathbf{H}$, either by expressed in hard tokens' way (e.g., task descriptions~\cite{vaswani2017attention} and task-related words~\cite{shin2020autoprompt}) or in soft tokens' fashion (e.g., task-based soft embeddings~\cite{qin2021learning,lester2021power}). Following the text-to-text paradigm, generally, these prompt methods convert downstream tasks into some cloze-style objectives. 

Take classification as an example. 

Given an input sentence $x$ and the label $y$, where $(x, y) \in \mathcal{D}$ where $\mathcal{D}$ refers to the dataset, we build a sequence of learnable soft prompts $h$, and we optimize the model with the cross-entropy loss as shown below:
\begin{equation}
\label{generalEq}
\mathcal{L} = -\sum_{ (x, y) \in \mathcal{D} } \log p(y | h, x).
\end{equation}
During the training process, we only update the parameters of $h$ and keep the PLM fixed. 

\para{Instance-specific Information Matters for Prompts.}
One important characteristic of existing methods outlined previously is the task-level prompt space: they assign a shared embedding (e.g., $\mathbf{H}$ in \ref{generalEq}) for all input data in a task, regardless of the specific content of each instance in the input. We suspect that a shared data-agnostic representation for all input may hinder PLMs' abilities by the difficiency in data-specific knowledge. 



To validate this conjecture, we experiment with the RoBERTa-large PLM and Prompt Tuning~\cite{lester2021power} to show the effect of changing the granularity of prompts from task level to instance level. We use Prompt Tuning as the baseline, and compare it against a very straightforward version of instance-wise prompts realized by the following changes.
We first initialize a token-level prompt table, instead of a single shared prompt, which maps each word in RoBERTa-large's vocabulary table to an embedding. The embeddings are randomly initialized. During training, we truncate instances by 100 tokens and add prompts simply by querying all the tokens in the prompt table. We keep all other settings identical with the vanilla Prompt Tuning (e.g., freezing PLM's parameters and tuning only prompts). We perform our experiments on two open benchmarks from SuperGLUE: BoolQ and CB. We represent vanilla Prompt Tuning as ``PT`` and the modified version as ``PT-v``. The results are reported in Figure.~\ref{observation}. 
The results show that instance-wise prompt tuning achieves visible performance improvement over Prompt tTning. We suspect this is because instance-wise prompts introduce extra parameters for word representations, which leads to a clear division of the prompt optimization space. To this end, embeddings of the prompt table can be optimized in a fine-grained way.


\para{Challenges.} We have shown the benefit of incorporating input-specific, instance-level information when querying PLMs for downstream tasks. However, the experiment only uses a very basic approach. Generally, making instance-wise prompt tuning more effective and efficient is non-trivial, with the following challenges to address. First, randomly initialized embeddings may be insufficient to extract enough background knowledge. Instance-wise prompts are supposed to implicate intrinsic knowledge behind specific words from input, such that they can better stimulate PLMs for downstream tasks. Second, the performance gain of instance-wise prompt tuning comes at the cost of more trainable parameters due to larger prompt embedding space. We expect to utilize instance-wise prompt tuning in an effective and efficient manner. 
We will next show how our technique reduces tuned parameters  of prompt embeddings without affecting the final performance.




%% file: 4.method.tex
\section{The proposed method}
\label{method}
In this section, we first introduce the formulation of the \sys technique. Then we introduce three instance-wise prompt strategies. We illustrate how we design these strategies and how we leverage pretraining to empower instance-wise prompts with intrinsic data knowledge, and how to apply them to specific tasks to improve task performance. 

\subsection{Formulation and Overview}
\para{\sys Formulation.}
\label{formulation}
\sys formulation is similar with general prompt methods' formulation proposed in Section \ref{formulation_general}, the only difference lies in the prompt format. Generally, \sys produces instance-wise prompts by a module called the \emph{IPT strategy}, denoted as $\mathbf{G_{\omega}}$ (parameterized by $\omega$).



Given a data instance $(x, y)$, the IPT strategy firstly generates a series of prompt tokens $t = \{t_1, t_2, \cdots, t_k\}$ for $x$ by $t = \mathbf{G_{\omega}}(x)$, where $k$ is a hyper-parameter that denotes the number of prompt tokens for each instance. Similarly, the prompt is then concatenated to the input data forming a single prompted instance representation $[t, e(x)]$ for downstream tasks, where $e(\cdot)$ refers to the embedding layer of the PLM. 
We minimize the cross-entropy loss by $\mathcal{L} = -\sum_{ (x, y) \in \mathcal{D} } \log p(y | t, x).$
The only tunable parameters during training are parameters of \sys strategy parameterized by $\omega$.


\para{IPT Module in the Full Pipeline.}
\begin{figure*}[htpb]
    \centering
    \includegraphics[width=.9\linewidth]{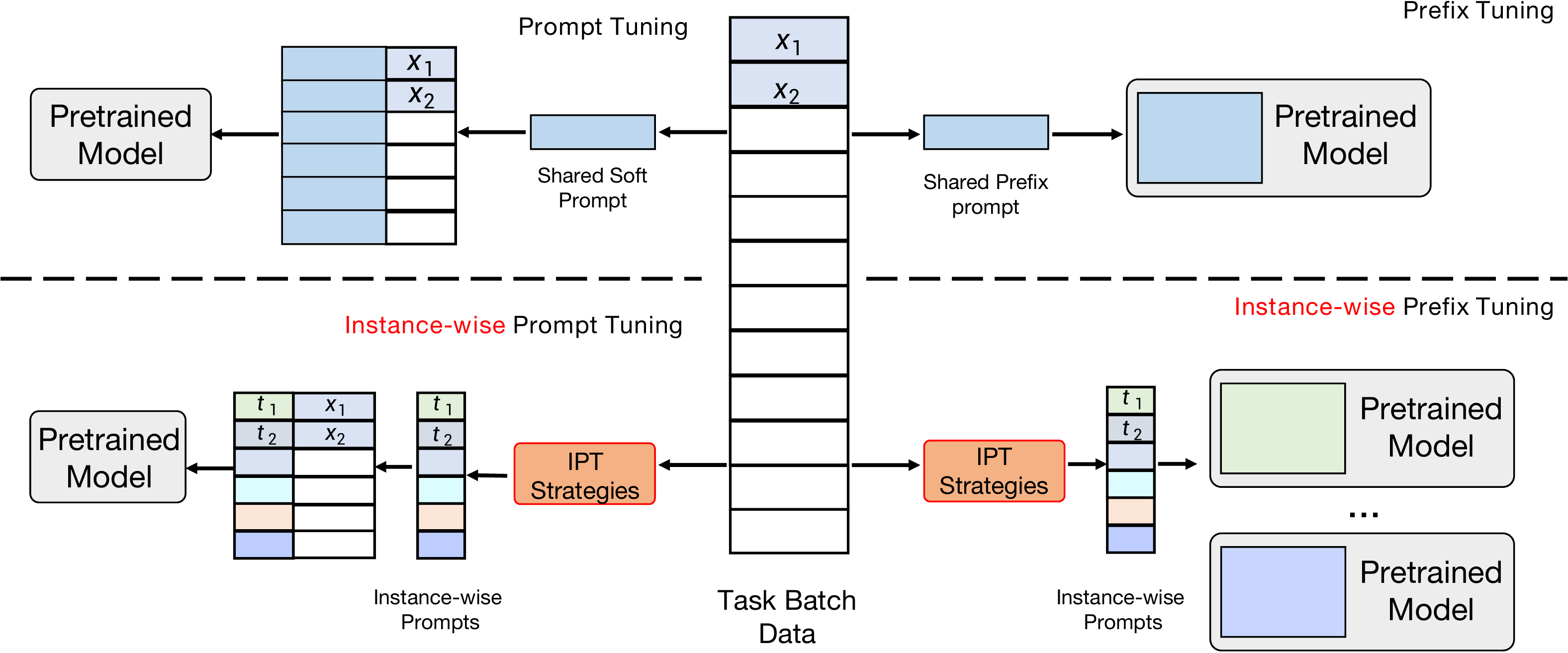}
    \caption{Pipeline of adopting \sys in major prefix tuning methods.}
    \label{pipeline}
    \vspace{-2mm}
\end{figure*}


We illustrate how \sys can develop its strength in the full prompt training process. We provide the examples of adopting \sys in popular families of Prefix tuning methods. We take Prompt Tuning and Prefix Tuning as base prompt models and plot how \sys module works in the training process in Figure. \ref{pipeline}. The upper part represents general processes of Prompt Tuning and Prefix Tuning respectively, and the underneath part of them draws how \sys can be appended to their pipelines. 


Following the task-based strategy, Prefix tuning methods allocate a shared soft embedding space according to different tasks. They directly concatenate the data-agnostic prompt to all input data and feed the reformed data to PLMs for downstream tasks. To improve the model performance with data-side knowledge, we propose three strategies to serve as the '\sys Strategies' part for different needs. 

\subsection{IPT Strategies}
In this section, we describe the three IPT strategies designed for different goals. We also show how to leverage pretraining to empower instance-wise prompts with intrinsic data knowledge, and how to apply them in specific tasks to improve task performance. 

\para{Random \sys.} We name the strategy applied in Section~\ref{motivation} Random \sys. 
Specifically, we randomly initialize a prompt  embedding table for each word in the vocabulary. To generate a prompt representation, we project the input sentence with the aforementioned embedding and truncate it by a specified prompt length of $L$, and we concatenate the sequence with the embeddings of the sentence as the input of transformer. 
We freeze the parameters from the pretrained model and only tune the newly built prompt embedding. 
In comparison with model tuning, IPT costs much less as only prompt embedding should be optimized, and compared with previous prompt tuning methods, Random \sys does not bring additional computational overhead, where the computation FLOPs (floating-point operations per second) of Random \sys and baselines are similar because of same tuned parameters in a batch. 




\para{Pretrained \sys.}
\label{pretrain-text}
Similar to Random \sys, Pretrained \sys has a prompt embedding table for tuning. However, we empirically assume that data-based knowledge can further benefit IPT. We note that existing methods use prompts to imply the task domains of the downstream tasks. Based on this insight, we expect to pretrain prompts with type knowledge from data, thus retrieving the knowledge of the  relevant category.



Data corpus in the real world is flooded with multiple type information, and they naturally implicate coarse-grained categorical knowledge. For example, a company email letter comes to stand for its type information of business activities. It is easy for humans to appreciate the type of information behind the language because human brains are skilled in reasoning. 
However, the data impedibility poses great challenges to language models for understanding extension behind data, because algorithms can hardly recognize data extension behind input text~\cite{miholca2018new}. 

\begin{figure}[tpb]
    \centering
    \includegraphics[width=1.\linewidth]{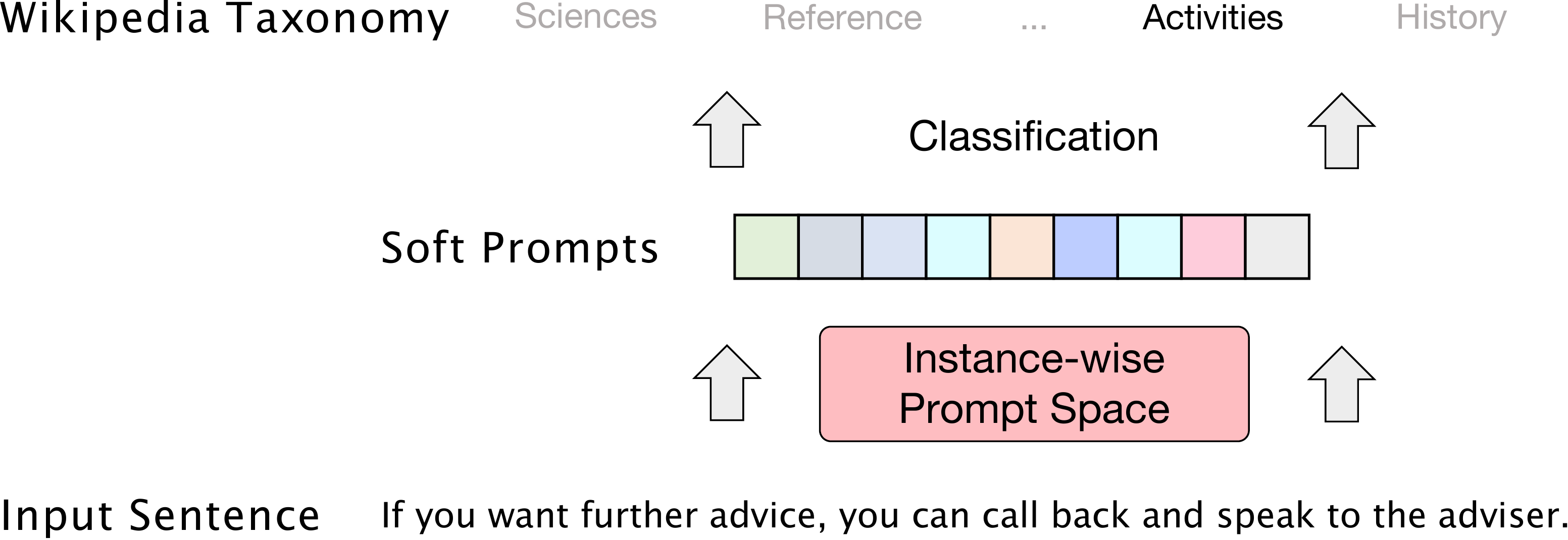}
    \caption{Data-driven Pretraining Process on the Instance-wise Prompt Space of \sys.}
    \vspace{-4mm}
    \label{pretrain}
\end{figure}

We develop a simple yet effective data-based pretraining policy to empower prompts with data-relevant knowledge by leveraging categorical information. 
For the data collection for pretraining \sys, we collect data from the Huggingface Datasets~\cite{HuggingFaceDataset}. 
To classify data corpus in a fine-grained and authoritative way, we follow Wikipedia's taxonomy~\cite{Wikipedia} and divide language corpus into $13$ categories, including society, technology, etc. 
However, tagging data with the classes should be time-consuming. 
Instead, we label data based on the original dataset description. 
To pretrain \sys, we build a simple text classifier and train it on the labeled data with cross-entropy loss $ \mathcal{L}_{p}= -\sum_{d_i \in \mathcal{D}} \mathbf{Y}_{ij} \log \mathbf{P}_{ij}$, 
where $\mathbf{Y}_i$ is a one-hot label indicator vector, and $\mathbf{P}_{i}$ is the model's predicted softmax outputs of the input data $d_i$. 
For more implementation details, please refer to the appendix A.3. 
After training the classifier, we extract the weights of the learned embedding for the initialization of our prompt embedding, which is different from random initialization in Random \sys. The pretraining process is represented in Figure. \ref{pretrain}.

\para{Encoder \sys}
\label{generator}
Note that both Random \sys and Pretrained \sys introduce embeddings, taking up roughly $1\%$ of parameters of RoBERTa-large. In order to reduce the computation costs, we introduce Encoder \sys. 


Similar to Pretrained \sys, we build a simple classifier with an embedding layer for representation extraction and a neural network, e.g., CNN, RNN, MLP, etc., for representation learning. 
Instead of extracting pretrained embedding layer as mentioned above, we retain the trained model for the construction of soft prompt. 
To be more specific, we fix the embedding layer and make the neural network tunable. 
Therefore, the amount of parameters to be optimized becomes smaller and requires fewer computation costs. 



For each input data $\mathbf{x}$ with $n$ tokens, we first determine a sequence utilization rate that controls the proportion of utilized tokens in the whole input text for the encoder. We feed the truncated representation of $\mathbf{x}$ into the encoder $\mathbf{G_{\omega}}$, and transform them to $\mathbf{t}$ with a length of $k$, where $\mathbf{t}= \mathbf{G_{\omega}}(\mathbf{x})$, $k < n$. $\mathbf{t}$ is a refined, data-specific and type-aware prompt for $\mathbf{x}$. 
Generally, fixing the prompt table and only tuning on $\omega$ can be efficient, in terms of both performance and cost. For example, Prefix Tuning of \sys version can outperform the original version by a large margin with only 0.5\% trainable parameters per task compared to fine-tuning, which counts a third of the original version's. 

Another important advantage brought from the prompt encoder architecture is its flexibility. 
The general encoder setting of \sys makes it pluggable that it could be substituted to all types of encoders, e.g., CNN encoders, RNN encoders, MLP encoders, etc. 
This feature endows \sys with a universality that it can easily extend to different domains with different encoders. 


\subsection{Discussions}
\begin{table*}[t]
\small
\centering
\caption{Overview of Current Prompt-based methods, including hard prompts and soft prompts. } \label{methods_all}
\resizebox{1.\linewidth}{!}{
\begin{tabular}{ccccccc}
\toprule
\textbf{Methods}&\textbf{Finetuned Params}& \textbf{Prompt Design}&\textbf{Few-shot}&\textbf{Middium-scale Model Support}&\textbf{Pretrain}&\textbf{Pretrain Pattern}\\
\midrule
GPT-3~\cite{vaswani2017attention}& None & Manual Designed Tokens &$\checkmark$ & $\times$ &$\times$& $\times$ \\
Prompt Tuning~\cite{lester2021power}& Task-based Prompt Embeds & Continuous Prefix &$\times$ & $\checkmark$ &$\times$ &$\times$\\
Prefix Tuning~\cite{li2021prefix}& Task-based Prompt Embeds & Continuous Prefix &$\checkmark$ & $\checkmark$ &$\times$ &$\times$\\
P-Tuning~\cite{liu2021gpt}& All + Task-based Prompt Embeds & Continuous Prompts &$\checkmark$ & $\checkmark$ & $\times$& $\times$\\
PPT~\cite{gu2021ppt}& Task-based Prompt Embeds & Continuous Prefix &$\checkmark$ & $\times$ & $\checkmark$& Task-based\\
SPoT~\cite{vu2021spot}& Task-based Prompt Embeds & Continuous Prefix & $\checkmark$ & $\checkmark$ & $\checkmark$& Task-based\\
\midrule
IPT& Instance-wise Prompt Embeds & Continuous Prefix & $\checkmark$ & $\checkmark$ & $\checkmark$& Data-based \\
\bottomrule
\end{tabular}}
\end{table*}

\para{\sys vs. general Prompt Learning methods.}
Prompt methods have gained much momentum for their superiority in tuning models in a lightweight way. 
However, most of them only work under constrained scenarios. For example, \cite{lester2021power} achieves comparable performance with finetuning when the model scale grows to more than $10$ billion in the full data regime, and
\cite{gu2021ppt} introduce prompt pretraining to address data constraints. 
Compared to these methods, \sys further relieves large-scale PLM restrictions and achieves competitive performance under different resource constraints. We summarize the characteristics of the existing prompt methods and \sys in Table \ref{methods_all}. 

Beyond universality, \sys also differs from previous works in prompt design. 
Previous works use fixed prompts for all data under a task. For example, some works (e.g., GPT-3) elect discrete and brief prompts like "Translate this sentence into English", and soft prompts works often optimize a shared and latent representation as prompts for downstream tasks. However, these methods hardly can be developed into an instance-wise manner. 

\para{\sys vs. Pretrained Prefix tuning methods.}
\cite{gu2021ppt,qin2021exploring} propose a series of prompt pretraining methods (PPT) to boost model performance across different tasks in the setups of few-shot and zero-shot learning. 
\cite{gu2021ppt} propose pretraining objectives based on the type of downstream tasks, and \cite{qin2021exploring} trains a meta-prompt on hundreds of language tasks to form a powerful prompt. One common ground of them is that they develop task-based pretraining policies. 
The idea of task-based prompt pretraining is orthogonal to IPT. 
In IPT, we pretrain instance-wise prompts with knowledge relevant to input data. 
It is available to pretrain prompts with both task-level and instance-level information. 
Besides, different from PPT, \sys rivals or outperforms all the baselines across tasks in both few-shot and full-shot settings. 
This demonstrates its robustness and flexibility under different resource limitations. 

%% file: 5.experiment.tex
\section{experiment}
\label{experiments}
We now empirically evaluate the effectiveness of \sys on open benchmarks. We aim to answer three questions as follows. \textbf{Q1}: Can \sys help prompt models achieve better performance by enabling instance-wise information? Can instance-wise prompts remain its superiority under different resource limitations? \textbf{Q2}: Why does the data-based knowledge helps when querying PLMs for downstream tasks? \textbf{Q3}: How do the hyperparameters influence the performance of \sys?

\subsection{Experiment Setup}
\para{Datasets.}
We study downstream performance on a diverse set of tasks from a widely acknowledged natural language understanding benchmark SuperGLUE~\cite{wang2019superglue}, including BoolQ~\cite{clark2019boolq}, CB~\cite{de2019commitmentbank}, RTE~\cite{dagan2005pascal} , MultiRC~\cite{khashabi2018looking}. Besides, we also evaluate \sys on open benchmarks of IMDB~\cite{imdb}, SST-2~\cite{sst} and AGNews ~\cite{agnews}. 

More details about the dataset description can be found in Section A.1 of the supplementary material. 


\para{Evaluation protocol.}
To avoid unstable results, we measure average performance of each experiment across 5 trials with different random seeds. We also observe that hyper-parameters can make a significant difference, and thus we sweep multiple hyper-parameters for each trial, and report the best result. 

For few-shot experiments, we follow previous few-shot works like \cite{gao2020making} to make a fair comparison. We use cross validation to perform principled model selection. Concretely, we sample 2K examples from each label from datasets and use 4-fold cross validation to determine the best hyperparameters. After finding the best hyperparameters, we train on the first K examples and early stop on the second K examples. We use K = 32 in this experiment and take Prefix Tuning as the base model. We perform the few-shot experiment on all four SuperGLUE datasets we used, and our examples are sampled from each dataset’s original training set. 

\para{Baselines.} 
Prompt Tuning~\cite{lester2021power} and Prefix Tuning~\cite{li2021prefix} are the two main baselines in our experiments. 
Also, we compare \sys with task-based prompt pretraining method PPT~\cite{gu2021ppt}. 

The optimal hyper-parameter settings may vary across the compared methods.
To enable more fair comparisons, we perform a hyper-parameter search on each model based on results on the validation set. 
The searching results of hyper-parameters and more details about baseline methods are provided in Section A.3 and Section A.4 in the supplementary material. 

\para{Implementations.}
We use OpenPrompt~\cite{openprompt} to implement \sys and baseline methods. Our code is available in the supplementary material. 

\para{Pretrained Language Model.}
We take RoBERTa-large as the backbone Pretrained Language Model in our experiments. 

\subsection{End-to-End Comparison}
To answer \textbf{Q1}, we implement different \sys strategies on both Prompt Tuning and Prefix Tuning. 

To examine the universality of \sys, we perform a thorough evaluation on NLU tasks in both full-shot and few-shot regimes. 

For a base model $\mathcal{M}$, we equip $\mathcal{M}$ with the proposed instance-wise prompt strategies. To evaluate the effectiveness of Encoder \sys, we adopt three typical encoders of different types, including LSTM~\cite{lstm}, TextCNN~\cite{textcnn} and MLP. 

\para{Full-data Regime.} We summarize the results on four datasets of SuperGLUE in a full data regime in Table \ref{end2end}. First, we could see \sys improves all base methods in all cases. The improvements over the baselines are up to 2.7\%, 10.8\%, 5.8\% and 9.4\% respectively on four datasets, which demonstrates the instance-wise information is significant for querying PLMs. 

Second, an inspiring observation is that \sys can be utilized to acquire better performance than fine-tuning in medium-scale PLMs. The results indicate that \sys relieves previous works' limitations of conditional advantages in specific scenerios (e.g., effective in large-scale model settings). This proves \sys's universality in bridging the gap between PLMs and downstream tasks across a wide range of NLU tasks. Besides, though Prefix Tuning achieves superior performance on natural language understanding tasks, by equipped with \sys's strategies, it can outperform fine-tuning in two of four datasets and become a new state-of-the-art method. This indicates that with the integration of advanced techniques, such as \sys, methods can be more competitive, and the superiority can be extended to multiple situations. This result demonstrates that \sys is also universally effective across different base models. 

Third, it can be observed that Pretrained \sys generally outperforms Random \sys. This illustrates that the data-based pretraining injects data knowledge to the instance-wise prompts, thus benefiting the downstream tasks. Besides, though with fewer tuned parameters, Encoder \sys achieve considerably, even the best performance in some cases, demonstrate instance-wise prompt tuning can be both effective and economic. 



Finally, it can be observed that the most effective instance-wise prompt strategies on these datasets are different. This illustrates the proposed strategies can develop their own advantages in different data distributions. 



\begin{table}[t]
\centering
{
\noindent
\caption{Results of end-to-end full-data comparisons on SuperGLUE datasets. We highlight the best performance with bold font, and the second best with underlines. }
\vspace{-4mm}
\label{end2end}
\renewcommand{\multirowsetup}{\centering}
\resizebox{1.\linewidth}{!}{
\begin{tabular}{c|c|c|c|c}
\toprule
\textbf{Models}& \textbf{BoolQ}&\textbf{CB}&\textbf{RTE}&\textbf{MultiRC}\\
\midrule
Prompt Tuning& $62$ & $71.3 $ & $50.2 $ & 56.7\\
PPT& 62.2  & 70.9 & 53.5 & 56.4  \\
Prefix Tuning&85& $\textbf{100.0}$& 84.8 & 82.5 \\
Fine-tuning &\textbf{86.9}&98.2& 86.6 & \textbf{85} \\
\midrule
Prompt Tuning + Random \sys &$64.2 $ &$75$& 54.5 & 62.9 \\
Prompt Tuning + Pretrained \sys &$64.6 $& $76.8$& 54.8 & 66.1\\
Prompt Tuning + Encoder \sys (CNN) &$64.7 $ &$\underline{82.1}$ & 56.0 & 59 \\
Prompt Tuning + Encoder \sys (RNN) &$ 64.5 $&$78.4$& 54.5 & 59.4 \\
Prompt Tuning + Encoder \sys (MLP) &$64.4 $ & $77.3$& 54.2 & 59.7 \\
\midrule
Prefix Tuning + Random \sys &$ 86.2 $&$\textbf{100.0}$ &$\textbf{88.8} $ & 83 \\
Prefix Tuning + Pretrained \sys &$ \underline{86.4} $& $\textbf{100.0}$ &\textbf{88.8} & \underline{83.5} \\
Prefix Tuning + Encoder \sys (CNN) &$85.4 $ & $\textbf{100.0}$ & \underline{86.2} & 83.3\\
Prefix Tuning + Encoder \sys (RNN) &$ 85.8$& $\textbf{100.0}$ & 85.9 & 83.1  \\
Prefix Tuning + Encoder \sys (MLP) & 85.9 & $\textbf{100.0}$ & 85.5 & 82.9 \\
\bottomrule
\end{tabular}}}
\vspace{-6mm}
\end{table}

\para{Few-shot Learning.} 
Table \ref{fewShot} illustrates how standard Finetuning and our \sys compare. \sys's strategies can outperform finetuning on three of four datasets in our experiments, demonstrating that \sys also capable of the gap between pretraining and fine-tuning in a few-shot scenerios. We can see that Pretrained \sys outperforms other \sys strategies on two of four datasets, the performance gain is brought by data knowledge implicated in prompts. Moreover, We have a similar observation in the full-data regime that no instance-wise prompt strategy demonstrates its superiority under all circumstances. For example, Pretrained \sys achieves the best performance on the BoolQ dataset, and the best-performing strategy of \sys on CB is the Random \sys. This is because both task content and data distributions of different benchmarks are diverse, making them  suitable for different kinds of instance-wise strategies. 

Besides, it can be observed that \sys can improve the base Prefix Tuning by a large margin of 11.2\%, 12.8\%, 2.9\% and 2.4\% respectively on the four datasets. This illustrates instance-wise prompts can implicate intrinsic knowledge thus improving the base model's ability to learn tasks with limited examples. The superior performance of \sys shows instance-wise knowledge helps in various data resource settings, indicating that \sys can be flexibly applied to diverse applications which have different data requirements. 

\begin{table}[t]
\caption{ Test accuracy (\%) in Few-shot learning scenarios. We find that IPT can generally outperform both finetuning and Prefix Tuning. }
\label{fewShot}
\centering
{
\noindent
\renewcommand{\multirowsetup}{\centering}
\resizebox{1.\linewidth}{!}{
\begin{tabular}{c|cccc}
\toprule
\textbf{Models}& \textbf{BoolQ}&\textbf{CB}&\textbf{RTE}&\textbf{MultiRC}\\
\midrule
finetuning & 51 & \textbf{74.3} & 54.5 & 54.8 \\
\midrule
Prefix Tuning & 54.9 & 57.1 & 53.7 & 57.1 \\
Prefix Tuning + Random \sys & 61.7 & 69.6 & 54.8 & \textbf{57.2} \\
Prefix Tuning + Pretrained \sys & \textbf{62.1} & 69.7 & 55.2 & \textbf{57.2}\\
Prefix Tuning + Encoder \sys (CNN) & 61.9 & 66.0 & 55.2 & 57.1\\
Prefix Tuning + Encoder \sys (RNN) & \textbf{62.1} & 67.8 & 54.5 & 57.1 \\
Prefix Tuning + Encoder \sys (MLP) & 62.0 & 67.8 & \textbf{56.6} & 57.1\\
\bottomrule
\end{tabular}}}
\vspace{-4mm}
\end{table}

\subsection{Interpretability}
To answer \textbf{Q2}, we here investigate why \sys is effective. Our interpretations on the performance gain is based on the type-related knowledge implicated in prompts. \sys introduces background type knowledge in the pretraining process, thus each input data in a task can query the PLM with the categories to extract its corresponding domain knowledge.


We visualize our instance-wise prompts produced by Pretrained \sys to illustrate its capability in supplying type-related knowledge. We randomly select 4000 unused sentences from different datasets, and represent them by PLM's embeddings and by our pretrained prompt embeddings respectively. 

\begin{figure}[tpb]
    \centering
    \includegraphics[width=1.\linewidth]{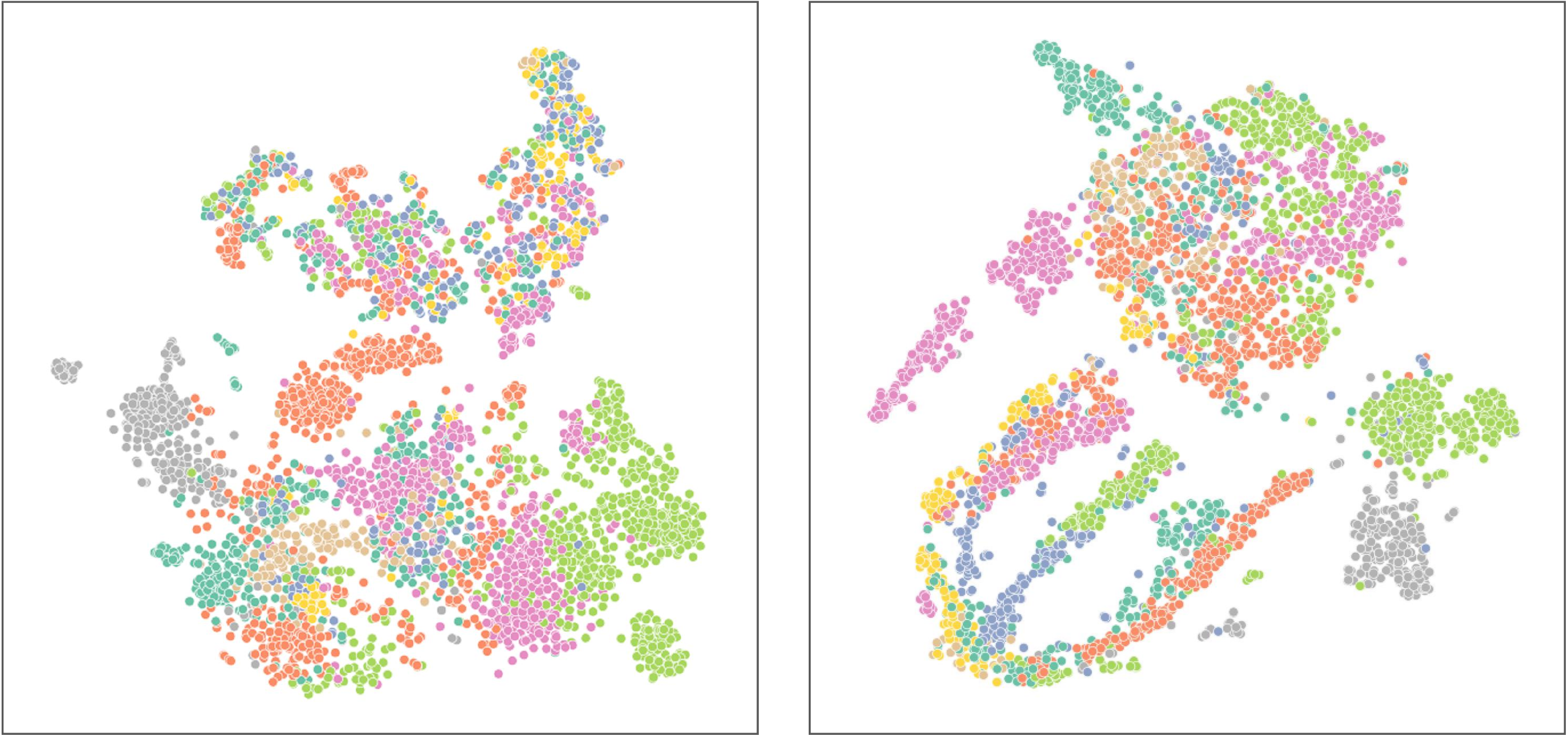}
    \vspace{-4mm}
    \caption{Sentence Representation Visualization.}
    \label{visualization}
    \vspace{-8mm}
\end{figure}

The left part in Figure.~\ref{visualization} is the sentence embeddings represented by the PLM. Benefiting from prior knowledge derived from the pretraining process, some homologous sentences (i.e., points of the same color) are close tightly. However, most of the sentence representations of different types are  mixed together. This indicates that PLMs naturally contain some knowledge that can help discriminate data under different contexts, but this is not enough to fully express the data divergence, as the data of different types are very close to each other. The right part of Figure.~\ref{visualization} visualize the instance-wise soft prompts of the input data. Compared with the left part, the data points are more scattering in the right part, and data belonging to the same type has an average shorter distance to each other. The visualization results indicate that instance-wise prompts implicate implicit data extension, thus retrieving PLM's knowledge with specific coarse-grained type knowledge which can boost the performance of downstream tasks.

\begin{figure*}[tb]
    \centering
    \includegraphics[width=.9\linewidth]{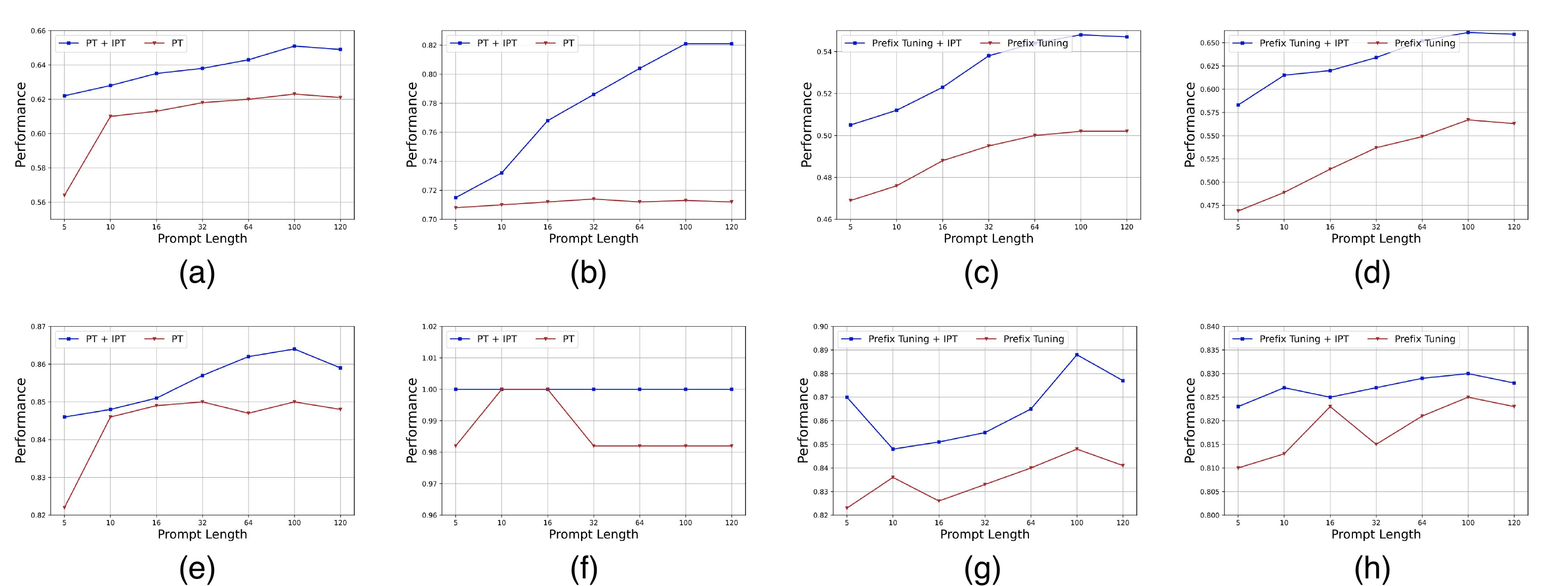}
    \vspace{-4mm}
    \caption{Prompt Length's Effects on \sys.}
    \label{prompt-length}
    \vspace{-4mm}
\end{figure*}

\subsection{Case studies}
The visualization part illustrates prompts of \sys implicate type-related knowledge. To further answer \textbf{Q2}, we illustrate why instance-wise prompts and categorical knowledge helps by case studies respectively. We provide qualitative examples from Boolq of the QA dataset and CB of the NLI dataset. 

We perform case studies on Random \sys and Encoder \sys, to evaluate the benefits brought by instance-wise prompts and type-related knowledge respectively. For each instance, we retrieve the prompt embeddings generated by trained Prompt Tuning, Random \sys and Encoder \sys, and calculate the distance between these prompts and the input tokens by calculating the cosine similarity distance between the generated prompt embeddings and token embeddings provided in PLMs. We color words which have the smallest distance to the prompt representations with red in Table.\ref{case}. We find that words selected by Prompt Tuning and Random \sys are hard to understand literally. This is because their prompt spaces are randomly initialized, thus the prompts can hardly implicate data-side information. Though hardly contain explicable information in prompts, by introducing more tunable parameters and optimizing in a fine-grained way, Random \sys still can predict right answers while Prompt Tuning gives wrongs. 

Besides, we find that Encoder \sys provides intuitive results. For example, we could infer that the instance is associated with the topic of science and technology by the provided words 'lunar' and 'astronauts' from the QA case. The result illustrates \sys apply instance-wise inner knowledge and type-informative knowledge to query PLMs, which may be credited for \sys superior performance on downstream tasks. 

\begin{table*}[t]
\caption{Case studies of \sys.}
\label{case}
\centering
{
\noindent
\renewcommand{\multirowsetup}{\centering}
\resizebox{1.\linewidth}{!}{
\begin{tabular}{ccccc}
\toprule
\textbf{Task}& \textbf{Input Instance}&\textbf{Prompt Tuning}&\textbf{Random \sys}&\textbf{Encoder \sys}\\
\midrule
QA & \makecell[c]{"question": "did we leave the lunar rover on the moon",\\ "passage": "Lunar Roving Vehicle -- The LRV was\\ transported to the Moon on the Apollo Lunar Module (LM)\\ and, once unpacked on the surface, could carry one\\or two astronauts, their equipment, and lunar samples. \\The three LRVs remain on the Moon.", "label": true } & \makecell[c]{$\mathbf{Predicted Label: false}$\\"question": "did we \red{leave} the lunar rover \red{on} the moon",\\ "passage": "Lunar Roving Vehicle -- The LRV was\\ transported to the Moon on the Apollo Lunar Module (LM)\\ and, once unpacked on the surface, could carry one\\or two astronauts, \red{their} \red{equipment}, and lunar \red{samples}. \\The three LRVs remain on the Moon."} & \makecell[c]{$\mathbf{Predicted Label: true}$\\"question": "did we leave the lunar \red{rover} on the moon",\\ "passage": "Lunar Roving Vehicle -- The LRV was\\ \red{transported} to the Moon on the Apollo Lunar Module (LM)\\ and, \red{once} unpacked on the surface, \red{could} carry one\\or two astronauts, their equipment, and lunar \red{samples}. \\The three LRVs remain on the Moon." } & \makecell[c]{$\mathbf{Predicted Label: true}$\\"question": "did we leave the lunar rover on the moon",\\ "passage": "\red{Lunar} Roving Vehicle -- The LRV was\\ transported to the \red{Moon} on the Apollo Lunar Module (LM)\\ and, once unpacked on the surface, could carry one\\or two \red{astronauts}, their equipment, and \red{lunar} samples. \\The three LRVs remain on the Moon." }\\
\midrule
NLI & \makecell[c]{"premise": "Most people are familiar with the idea of St. Bernards\\ or other dogs taking part in rescue and recovery efforts.\\ Robots might also take part in search and rescue missions.",\\ "hypothesis": "Robots are used to find missing victims.", \\"label": "entailment"} & \makecell[c]{$\mathbf{Predicted Label: contradiction}$\\"premise": "Most people are \red{familiar} with the idea of St. Bernards\\ or \red{other} dogs taking \red{part} in rescue and recovery efforts.\\ Robots might also take part in \red{search} and rescue \red{missions}.",\\ "hypothesis": "Robots are used to find missing victims."} & \makecell[c]{$\mathbf{Predicted Label: entailment}$\\"premise": "Most \red{people} are familiar with the idea of St. Bernards\\ or other dogs taking part in \red{rescue} and recovery efforts.\\ Robots might \red{also} take part in \red{search} and rescue missions.",\\ "hypothesis": "Robots are used to find missing \red{victims}."} &\makecell[c]{$\mathbf{Predicted Label: entailment}$\\"premise": "Most people are familiar with the idea of St. Bernards\\ or other dogs taking part in \red{rescue} and recovery efforts.\\ \red{Robots} might also take part in search and \red{rescue} missions.",\\ "hypothesis": "\red{Robots} are used to find missing victims."}\\
\bottomrule
\end{tabular}}}
\end{table*}

\subsection{Effectiveness of Knowledge-enhanced Pretraining}
We provide intuitive explains for why \sys is effective by visualization and case studies in the previous experiments. Now we answer $Q2$ by experimentally examining how type-related knowledge works from pretraining process. 

To justify whether type knowledge of instances can help query PLMs thus boost downstream tasks, we evaluate the proposed Pretraind \sys with Prefix Tuning on three open benchmark datasets, including AGNews, SST-2 and imdb respectively. We also experiment with vanilla Random \sys and a Random \sys variant. The variant appends hard tokens which implicates type knowledge of input data to the prefix of the Random \sys prompts. We utilize the variant to further validate our claim since type-related knowledge takes effects on querying PLMs for downstream tasks. Take both descriptions of datasets and Wikipedia's taxonomy, we allocate hard prompts of 'Human Activities' to AGNews, and allocate 'Culture and the Arts' to IMDB and SST-2. We report the results in Table. \ref{effectiveness}. As can be seen, the Random \sys variant with type-related prompt tokens outperforms plain Random \sys on all three datasets, demonstrating that type-related knowledge helps for PLMs querying. Furthermore, it can be observed Pretrained \sys can outperform the Random \sys variant on all compared datasets. This is because Pretrained \sys implicates the type knowledge in a soft format. Soft prompts can emphasize particular words (by lengthening their vectors) or particular dimensions of those words. In this way, prompts of \sys can imply type knowledge in a fine-grained way, and further boost the downstream tasks. 


\begin{table}[t]
\caption{Evaluation on the effects of type-related knowledge on prompting. }
\vspace{-2mm}
\centering
{
\noindent
\renewcommand{\multirowsetup}{\centering}
\resizebox{.8\linewidth}{!}{
\begin{tabular}{c|ccc}
\toprule
\textbf{Model}& \textbf{AGNews}&\textbf{SST-2}&\textbf{imdb}\\
\midrule
Random \sys & 86.2 & 96.0 & 95.9 \\
Random \sys + hard prompts & 86.7 & 96.1 & 96.1 \\
Pretrained \sys & 86.9 & 96.4 & 96.3 \\
\bottomrule
\end{tabular}}}
\vspace{-4mm}
\label{effectiveness}
\end{table}

\subsection{Hyperparameter studies}
In the following experiment, we provide results for various values of \sys's important hyper-parameters.

\para{Prompt Length} We train instance-wise prompts for each model size while varying the prompt lengths in ${5, 10, 16, 32, 64, 100, 120}$ and fixing other settings to our default configuration. To fully understand \sys's influence on the base methods under settings of different prompt lengths, we also report the effects of prompt lengths on the base methods. 
To be more specific, we adopt Pretrained \sys strategy for comparison. 
We demonstrate the experimental results on Figure~\ref{prompt-length}. We equip \sys with Prompt Tuning, and report the corresponding results on the four SuperGLUE datasets in Figure \ref{prompt-length} (a) - (d). We equip \sys with Prefix Tuning, and the results are reported in Figure \ref{prompt-length} (e) - (h). We have analogous findings with \cite{lester2021power}. We find that increasing the prompt length yields gains when the prompt length is shorter than $100$ words. 
Going past $100$ tokens appears to be mildly detrimental. This indicates that longer prompts can not always bring gains. Lengthy prompts query PLMs with noises, thus hindering the model performance on downstream tasks. The result also shows that CB is not sensitive to prompt length in Prompt Tuning, while the performance is positively correlated with the prompt length when \sys is added to the base model. This indicates that by increasing prompt length, the extended type knowledge behind instances progressively brings out data intrinsic knowledge and bridges the gap between the PLM and downstream tasks. Besides, it can be observed that \sys outperforms the baseline model by a large margin up to 6\% when the prompt length is set to a small value (e.g., 5). Hence we believe this indicates that, although prompts of this length includes little type knowledge instructions, the supplementary categorical knowledge from instance-wise prompts contain significant context to query the PLM. 

\para{Sequence Utilization Rate.}
In this part, we investigate how the sequence utilization rate of the input instance affects downstream performance. We introduce these parameters in the Encoder \sys part. Sequence utilization rate refers to the proportion of utilized tokens in the whole input sentence for instance-wise prompt generation. This is an interesting and considering question for Encoder \sys because we can further understand the data integrity's effects on the instance-wise prompt process. 

We take Prompt Tuning as the base model, and TextCNN, LSTM and MLP as the encoder component respectively. 
We keep the prompt length to $100$, and perform our experiments on the BoolQ dataset. We report the effects of sequence utilization rate in Table~\ref{encoder}. 
It can be observed that when the sequence utilization rate is set to a very small value (e.g., $0.2\%$), the instance-wise knowledge contributes little to downstream tasks.  With the increase in the sequence utilization rate, more data-related knowledge is injected into the prompts, and the performance on all encoders increase correspondingly. Besides, when the sequence utilization rate grows up to more than $26\%$, the performance of all encoders appear to be stable. The result illustrates that our algorithms can successfully tell apart instance differences by front parts of the input. 
Increasing the sequence utilization rate can hardly bring significant marginal benefits when its numerical value has already set large. 


\begin{table}[t]
\caption{Evaluation of sequence utilization rate's effects on the quality of generating instance-wise prompts. }
\centering
{
\noindent
\renewcommand{\multirowsetup}{\centering}
\resizebox{1.\linewidth}{!}{
\begin{tabular}{c|cccccccc}
\toprule
\textbf{Sequence Utilization Rate}& \textbf{0.2\%}&\textbf{1.3\%}&\textbf{3\%}&\textbf{5\%}&\textbf{10\%}&\textbf{21\%}&\textbf{26\%}&\textbf{30\%}\\
\midrule
Encoder \sys (RNN)& 60.3 & 61.4 & 61.5 & 62 & 62.8& 63.4 & 64& 64\\
Encoder \sys (CNN)& 61.2 & 61.6 & 62.1 & 62.4 & 63.4 & 64.6& 64.7& 64.6\\
Encoder \sys (MLP)& 60.7 & 61.9 & 62.3 & 62.7 & 63.3& 64.4& 64.3& 64.3\\
\bottomrule
\end{tabular}}}
\vspace{-4mm}
\label{encoder}
\end{table}

%% file: 6.appendix.tex
\appendix
\section{Outline}
The appendix is organized as follows:
\begin{description}
    \item[A.1] Datasets description.
    \item[A.2] More details about the compared baselines.
    \item[A.3] Experimental settings.
    \item[A.4] Experimental environment and reproduction instructions.
\end{description}

\subsection{Dataset Description.}
\label{dataset_description}
\noindent\textbf{BoolQ} ~\cite{clark2019boolq} is a QA task where each example consists of a short passage and a yes/no question about the passage. The questions are provided anonymously and unsolicited by users of the Google search engine, and afterward paired with a paragraph from a Wikipedia article containing the answer.

\noindent\textbf{CB} ~\cite{dagan2005pascal}  is a corpus of short texts in which at least one
sentence contains an embedded clause. Each of these embedded clauses is annotated with the degree
to which it appears the person who wrote the text is committed to the truth of the clause. The resulting
task is framed as three-class textual entailment on examples that are drawn from the Wall Street Journal,
fiction from the British National Corpus, and Switchboard. Each example consists of a premise
containing an embedded clause and the corresponding hypothesis is the extraction of that clause.

\noindent \textbf{RTE} ~\cite{de2019commitmentbank}  datasets come from a series of textual entailment challenges. Examples are constructed based on news and Wikipedia text. All datasets are combined and converted to two-class classification: entailment and not entailment. 

\noindent\textbf{MultiRC} ~\cite{khashabi2018looking}  is a QA task where each
example consists of a context paragraph, a question about that paragraph, and a list of possible
answers. The system must predict which answers are true and which are false. The
paragraphs are drawn from seven domains including news, fiction, and historical text.

\noindent\textbf{IMDB} ~\cite{imdb}  is a sentiment classification dataset about movie reviews, which has two classes.

\noindent\textbf{SST-2} ~\cite{sst}  is a sentiment classification dataset about movie reviews, which has two classes.

\noindent\textbf{AGNews} ~\cite{agnews}   is a news  topic classification dataset, gathered from news sources by ComeToMyHead in more than 1 year of activity. ComeToMyHead is an academic news search engine which has been running since July, 2004.

\subsection{Compared Baselines.}
The main characteristic of all baselines are listed below:
\label{baseline_methods}
\begin{itemize}
    \item Finetuning: Fine-tuning means taking weights of a trained language model and use it as initialization for a new model ,with all parameters being trained on new data.
    \item Prompt Tuning~\cite{lester2021power}: Prompt Tuning introduces trainable continuous prompts to input layer as a substitution to natural language prompts for NLU when backbone language models’ parameters are frozen.
    \item Prefix Tuning~\cite{li2021prefix}: Prefix Tuning directly prepend a series of continuous prompts to each layer of the based PLMs, while keeping the PLM’s parameters frozen. Prefix-tuning was originally proposed for natural language generation (NLG) tasks.
    \item PPT~\cite{gu2021ppt}: PPT propose to pretrain prompts by adding soft prompts into the pre-training stage to obtain a better initialization.
\end{itemize}

\subsection{Experimental Settings.}
\label{hyperparameter_setting}
\paragraph{Hyperparameters}
In full shot task and few shot task, the batch size is set to 32 and the learning rate is set to 1.0e-5. We set the gradient accumulation steps 2 and num warmup steps 2000 .The prompt length varies from 1–120. Prompt Tuning and  Prefix Tuning  is  trained for 20-100 epochs in full shot task while 100-800 epochs in few shot task. For the few shot task, following ~\cite{schick2020exploiting,schick2020s}, we randomly select 32 samples to construct the training set $D_{train}$ from the original training data and keep the samples across labels balanced. We  follow ~\cite{gao2020making} to select a validation set $D_{dev}$ from the original training data, which ensures that $|D_{train}| = |D_{dev}|$ and to use the original validation set as the test set $D_{test}$, which means $|D_{test}| >> |D_{train}| = |D_{dev}|$.

\paragraph{Pretrained IPT}
We train  a textcnn classification model for the sentence classification tasks. Following Wikipedia's taxonomy, we categorize datasets into 13 categories, including General reference, Culture and the arts, Geography and places, Health and fitness, History and events, Human activities, Mathematics and logic, Natural and physical sciences, People and self, Philosophy and thinking, Religion and belief systems, Society and social sciences and Technology and applied sciences. We acquire the pretraining data of these types from Huggingface Datasets~\cite{HuggingFaceDataset}. 

After training the classifier, we extract the weights of the learned embedding for the initialization of our prompt embedding, which is different from random initialization in Random IPT.

\paragraph{Encoder IPT}
We examine Encoder \sys's effects on three encoder networks, including TextCNN, MLP and LSTM. The TextCNN is composed of three convolutional layers and three pooling layers. The MLP network in our settings is composed of three liner layers. The LSTM network is composed of three LSTM layers and  three liner layers.
Every  encoder has only 0.5\% of pretrain model parameters.

\subsection{ Experiment Environment}
We run our experiments on a machine with 8 Tesla V100  GPUs.
The code is written in Python 3.7. We use Pytorch 1.9.0 on CUDA 10.2 to train the model on GPU.
